\documentclass[11pt]{article}

\usepackage[final]{acl}

\usepackage{times}
\usepackage{latexsym}

\usepackage[T1]{fontenc}

\usepackage[utf8]{inputenc}

\usepackage{microtype}

\usepackage{inconsolata}

\usepackage{graphicx}

\usepackage{amsmath} 
\usepackage{amssymb} 
\usepackage{booktabs}
\usepackage{multirow}
\usepackage{enumitem}
\usepackage{hyperref}
\usepackage{xurl}
\usepackage{amsthm}

\usepackage{listings}
\usepackage{xcolor}
\usepackage{array}

\definecolor{sepseq_green}{HTML}{0B6E4F} 
\definecolor{vanilla_grey}{HTML}{6E6E6E} 

\lstdefinelanguage{json}{
    string=[s]{"}{"},
    stringstyle=\color{-},
    comment=[l]{//},
    morecomment=[s]{/*}{*/},
    commentstyle=\color{gray}\itshape,
    numbers=left,
    numberstyle=\tiny,
    stepnumber=1,
    numbersep=8pt,
    showstringspaces=false,
    breaklines=true,
    frame=lines,
    backgroundcolor=\color{gray!10},
    literate=
     *{0}{{{\color{+}0}}}{1}
      {1}{{{\color{+}1}}}{1}
      {2}{{{\color{+}2}}}{1}
      {3}{{{\color{+}3}}}{1}
      {4}{{{\color{+}4}}}{1}
      {5}{{{\color{+}5}}}{1}
      {6}{{{\color{+}6}}}{1}
      {7}{{{\color{+}7}}}{1}
      {8}{{{\color{+}8}}}{1}
      {9}{{{\color{+}9}}}{1}
      {:}{{{\color{black}{:}}}}{1}
      {,}{{{\color{black}{,}}}}{1}
      {\{}{{{\color{black}{\{}}}}{1}
      {\}}{{{\color{black}{\}}}}}{1}
      {[}{{{\color{black}{[}}}}{1}
      {]}{{{\color{black}{]}}}}{1},
}

\newcommand{\ie}{\emph{i.e., }}
\newcommand{\eg}{\emph{e.g., }}

\definecolor{+}{RGB}{0,128,0} 
\definecolor{-}{RGB}{139,0,0}

\newtheorem{theorem}{Theorem}
\newtheorem{assumption}[theorem]{Assumption}

\title{SepSeq: A Training-Free Framework for Long Numerical Sequence Processing in LLMs}

\author{
    Jie Sun$^{1,2}$\hspace{0.5mm},
    Yu Liu$^{3}$\thanks{Project Lead}\hspace{0.5mm},
    Lu Han$^3$\hspace{0.5mm},
    Qiwen Deng$^4$\hspace{0.5mm}, \\
    \textbf{Xiang Shu$^5$\hspace{0.5mm},}
    \textbf{Yang Xiao$^6$\hspace{0.5mm},}
    \textbf{Xingyu Lu$^3$\hspace{0.5mm},}
    \textbf{Jun Zhou$^7$\hspace{0.5mm},} \\
    \textbf{Pengfei Liu}$^2$\hspace{0.5mm}
    \textbf{Lintao Ma$^8$\thanks{Corresponding authors}\hspace{0.5mm},} 
    \textbf{Jiancan Wu$^1$\footnotemark[\value{footnote}]\hspace{0.5mm},}
    \textbf{Xiang Wang$^1$\footnotemark[\value{footnote}]\hspace{0.5mm}} \\
    $^1$ University of Science and Technology of China \quad $^2$ Shanghai Innovation Institute \\
    $^3$ Nanjing University \quad $^4$ University of Edinburgh \quad $^5$ East China Normal University \\
    $^6$ Hong Kong Polytechnic University \quad $^7$ Zhejiang University \quad $^8$ Ocean University of China
}

\begin{document}
\maketitle

\begin{abstract}
While transformer-based Large Language Models (LLMs) theoretically support massive context windows, they suffer from severe performance degradation when processing long numerical sequences. 
We attribute this failure to the attention dispersion in the Softmax mechanism, which prevents the model from concentrating attention. 
To overcome this, we propose \textbf{Sep}arate \textbf{Seq}uence (\textbf{SepSeq}), a training-free, plug-and-play framework to mitigate dispersion by strategically inserting separator tokens.
Mechanistically, we demonstrate that separator tokens act as an attention sink, recalibrating attention to focus on local segments while preserving global context.
Extensive evaluations on 9 widely-adopted LLMs confirm the effectiveness of our approach: SepSeq yields an average relative accuracy improvement of 35.6\% across diverse domains while reducing total inference token consumption by 16.4\% on average.
\end{abstract}





\section{Introduction}
Transformer-based Large Language Models (LLMs)~\citep{transformers} now support increasingly long context windows~\citep{bg_llm_claude,bg_llm_gemini2.5pro,bg_llm_llama4,bg_llm_glm4.5,bg_llm_k2}, yet they remain unreliable on long numerical sequence processing tasks. 
In these tasks, the input is a sequence of numbers, and the model must answer a natural-language question that requires precise access to the sequence, such as counting, filtering, trend identification, or forecasting-style analysis. Unlike open-ended long-context understanding, these tasks are precision-critical: even a small error in preserving or attending to the sequence can change the final answer substantially.

This limitation is not resolved simply by extending the nominal context window~\citep {long_context_not_analyze_flawless,long_content_performance_drop}. Prior work has shown that effective context utilization often lags behind the advertised context length, especially when precise retrieval from long inputs is required. 
The challenge is particularly acute for numerical sequences, where many tokens are locally similar, semantically lightweight in isolation, and individually important to the final answer. While external tools~\cite{bg_tool_survey,bg_retool,bg_toolrl,bg_task_tool_xlam} or Program-of-Thought (PoT)~\cite{PoT} prompting can assist downstream computation, they still depend on the model's ability to faithfully preserve and reference the original sequence, leaving the upstream input-processing bottleneck largely unchanged.

\begin{figure*}[t]
    \centering
    \includegraphics[width=1.0\textwidth]{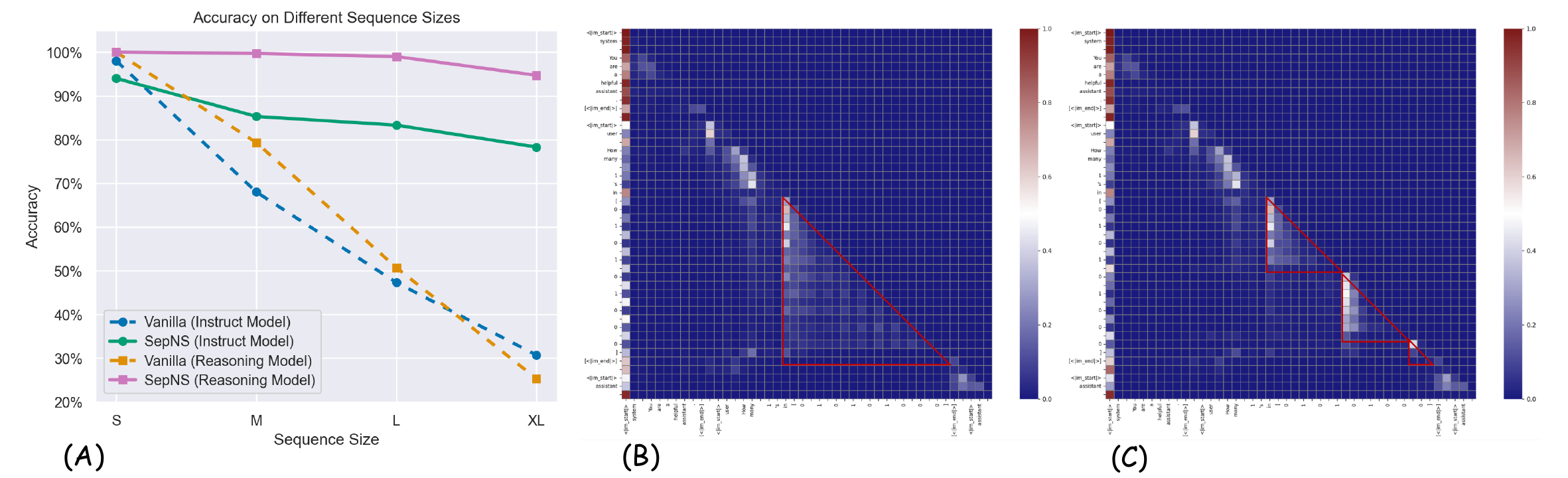}
    \caption{(A) Average accuracy across six synthetic tasks, performance drops sharply with increasing numerical sequence length (S: 2–32, M: 33–128, L: 129–256, XL: 257–512), while SepSeq remains largely unaffected.
    (B) visualize the attention scores given the input: ``... How many 1's in [0 1 0 1 0 1 0 0 0] ...''. (C) show the attention scores for the input with segmentation: ``... How many 1's in [0 1 0 1\textbackslash n0 1 0 0\textbackslash n0] ...''. (B) exhibits dispersed attention across the entire sequence while (C) demonstrates segment-focused attention.}
    \label{fig:motivation}
\end{figure*}

In this work, we identify attention dispersion as a practical bottleneck for long numerical sequence processing in LLMs. As the input grows longer, attention mass is distributed across a large number of numerically similar tokens, making it harder for the model to maintain focused access to locally relevant values~\cite{scalable_softmax,softmax_not_enough}. To mitigate this issue, we propose \textbf{Sep}arate \textbf{Seq}uence (\textbf{SepSeq}), a training-free and plug-and-play input-formatting framework that inserts separator tokens at regular intervals to partition a long numerical sequence into shorter segments. 

Although separator tokens have been employed in inference acceleration~\cite{SepLLM} and vision transformers~\cite{ViT}, their utility for training-free long-content question answering remains unexplored.
Functionally, SepSeq operates by introducing systematic separator insertion to define distinct boundaries, effectively transforming intractable long inputs into segments aligned with the model's effective capacity. This segmentation mitigates the precision degradation highlighted in Figure~\ref{fig:motivation}(A). Mechanistically, we show that the inserted separators act as attention sinks that attract subsequent attention and reduce cross-segment interference. This creates a more localized attention pattern, allowing the model to focus on shorter numerical spans while still preserving access to the full sequence, triggering specific heads to concentrate on local segments shown in Figure~\ref{fig:motivation}(B,C). Rather than introducing a new attention operator, SepSeq operationalizes a simple inference-time redistribution of attention that is particularly effective for precision-critical numerical inputs.

We conduct comprehensive experiments across 9 high-performance LLMs, evaluating performance on 6 synthetic tasks and 4 real numerical sequence processing domains.
Our results demonstrate that SepSeq substantially outperforms baselines, achieving significant average accuracy gains of 35.6\% across all evaluated datasets.
Notably, these performance gains are achieved with reduced computational overhead: the method requires no additional training and reduces token consumption by 16.4\% during inference.

In summary, our main contributions are as follows: 
\textbf{First}, we identify attention dispersion as the bottleneck limiting LLMs' performance on long numerical sequences and visualize how standard Softmax mechanisms fail to produce a sufficiently sharp attention distribution.
\textbf{Second}, we introduce SepSeq, a plug-and-play, training-free framework that overcomes this limitation by strategically inserting separators to recalibrate attention weights.
\textbf{Third}, we provide comprehensive empirical validation across diverse datasets, demonstrating that simple input formatting can unlock substantial performance gains (Avg. +35.6\%) and reduce 16.4\% token consumption during inference.

\section{Related Work}

\textbf{Long Context Modeling and Efficiency.} 
While methods like RoPE Scaling~\citep{scaling_RoPE} and YaRN~\citep{YaRN} have expanded context windows, mere architectural extension does not ensure robust utilization. Recent efficiency-focused works (\eg StreamingLLM~\citep{StreamLLM}, SepLLM~\citep{SepLLM}) address memory bottlenecks but do not fundamentally solve the reasoning degradation. This is evidenced by the ``Lost in the Middle'' phenomenon~\citep{liu2024lost} and the RULER benchmark~\citep{ruler}, which reveal that effective context length often lags far behind nominal window size, particularly for precision-critical tasks.

\textbf{Representation Bottlenecks in Numerical Processing.} 
Numerical processing presents unique challenges due to the discrete nature of tokenization contrasting with the continuous nature of numerical values. 
Crucially, recent theoretical analysis~\citep{transformers_need_glasses} suggests that long numerical sequences are prone to ``over-squashing'', where information is excessively compressed into fixed-size vectors. 
This compression leads to indistinguishable encodings, preventing the model from resolving specific values within dense sequences. Consequently, models frequently exhibit reasoning failures~\citep{li2025numeric_bench,hendrycks2021measuring}. 
While vocabulary adaptation via retraining presents a viable solution, the extensive requirements for high-quality data, computational resources, and specialized training techniques prove prohibitive for modern large-scale LLMs~\citep{kudo2018sentencepiece}. This bottleneck necessitates the development of lightweight, inference-time solutions.

\textbf{The Transcription Barrier in Tool Use.} 
A prevailing strategy to mitigate calculation errors involves leveraging external tools~\citep{schick2023toolformer, chen2022program, gao2023pal} or autonomous agents~\citep{yao2022react, shinn2023reflexion}. 
However, this paradigm merely shifts the burden from computational reasoning to textual transcription. 
Existing studies~\citep{welleck2019neural, zhang2022chinese} indicate that LLMs struggle to faithfully transcribe long sequences without corruption. 
In the context of long numerical processing, this limitation manifests as an inability to correctly formulate function calls with the complete, unaltered sequence. 
Consequently, the efficacy of tool-based or agentic methods remains fundamentally constrained by the model's intrinsic capability to attend to and preserve the raw input sequence.

\section{Method}
\label{sec:method}

In this section, we begin by establishing the formal problem setting for long numerical sequence processing.
Next, we analyze the inherent limitations of current LLMs to pinpoint the structural causes of their suboptimal performance. 
Finally, we introduce \textbf{Sep}arate \textbf{Seq}uence (\textbf{SepSeq}), a training-free, plug-and-play framework designed to optimize performance via strategic separator insertion.

\subsection{Problem Definition}
We study long numerical sequence processing tasks. Given a numerical sequence of length $n$
\begin{equation}
    S = [x_1, x_2, \ldots, x_n],
\end{equation}
where each $x_i$ is a numerical value, and a natural-language query $q$, the goal is to generate an answer
\begin{equation}
y = f_\theta(q, S).
\end{equation}

The query may require counting, comparison, filtering, aggregation, trend identification, or other reasoning operations over the sequence.
These tasks are challenging for three reasons. 
First, they are \textit{sequence-complete}: the answer may depend on any element in the sequence, so omitting or corrupting even a small part of the input can change the result~\citep{liu2024lost,faith_fate}. 
Second, they require \textit{position-sensitive retrieval}: the model must correctly access specific values or local spans within a long sequence rather than rely on coarse semantic summaries. 
Third, they are \textit{precision-critical}: many tasks involve exact counting or comparison, where small upstream errors can propagate directly into the final answer.


For example, given a stock-price sequence, the query ``Excluding non-trading days, how many times did the opening price rise for three or more consecutive days?'' requires the model to preserve the sequence faithfully, identify valid local patterns, and aggregate them under explicit constraints. This is not merely a language-understanding problem; it is a precision-critical interaction between natural-language instructions and long numerical input.


\subsection{The Bottleneck of Long Numerical Sequences}
Long numerical sequences are especially difficult for LLMs because they provide little semantic redundancy. In ordinary text, approximate semantic matching may still preserve the gist of the input. In contrast, numerical sequence tasks often require exact access to individual values or short local spans, and neighboring numbers can be superficially similar while playing very different roles in the final computation. As a result, small errors in attending to, preserving, or referencing the sequence can directly cause answer errors.

One may hope to address this challenge through tool use~\citep{bg_tool_survey}, for example, by asking the model to write code or invoke an external solver. However, this does not eliminate the core difficulty. Before any external computation can help, the model must still preserve the input sequence well enough to transcribe, reference, or pass it correctly to the tool. In long numerical inputs, this upstream requirement is itself fragile: if the model drops, alters, or misaligns even a small portion of the sequence, the subsequent tool execution can no longer be trusted. Therefore, tool-assisted reasoning is complementary to, but does not remove, the need for accurate long-sequence input processing.

Existing empirical studies~\citep{repetition-llm,pimentel2025repetitions,authors2025understanding_repetitions,transformers_need_glasses} highlight the susceptibility of LLMs to errors when transcribing long sequences. To quantify this, we evaluate the LLMs' ability to transcribe numerical sequences verbatim. As detailed in Appendix~\ref{apd:repetition-dilemma}, we observe a drastic performance collapse: for sequences exceeding 256 floating-point numbers, the success rate plummets to 14\%. This inability to preserve sequence integrity confirms that the bottleneck is intrinsic to the model's processing capacity, effectively rendering tool-based solutions ineffective.

We attribute this bottleneck to attention dispersion in the standard Softmax attention mechanism:
\begin{equation}
    \text{Attention}(\mathbf{Q}, \mathbf{K}, \mathbf{V}) = \text{softmax}\left(\frac{\mathbf{Q}\mathbf{K}^T}{\sqrt{d_k}}\right) \mathbf{V},
\end{equation}
where $\mathbf{Q}, \mathbf{K}, \mathbf{V} \in \mathbb{R}^{N \times d_k}$ denote the query, key, and value matrices, respectively, with $N$ representing the sequence length.
The Softmax function normalizes the attention scores to ensure the weights sum to 1: $\alpha_i = \frac{\exp(s_i)}{\sum_{j=1}^N \exp(s_j)}$. Here, $s_i$ denotes the scaled dot-product logit, specifically the element of $\frac{\mathbf{Q} \mathbf{K}^T}{\sqrt{d_k}}$ corresponding to the alignment between the query and the $i$-th key.

As the sequence length increases, the normalization denominator accumulates contributions from many irrelevant or weakly relevant tokens. For long numerical inputs, where tokens are numerous and often locally similar, this effect dilutes the attention allocated to the truly relevant values. Consequently, the model's attention becomes less focused and more diffuse, making it harder to retrieve the exact numerical evidence needed for downstream reasoning. This phenomenon, illustrated in Figure~\ref{fig:motivation}(B), often termed attention dispersion~\citep{attention_dispersion_HICD,attention_dispersion_MateICL,attention_dispersion_EMA}, causes the model to lose focus on critical context, leading to the failures observed in Figure\ref{fig:motivation}(A).

\subsection{SepSeq: Separate Sequence}

\begin{figure*}[t]
    \centering
    \includegraphics[width=1.0\textwidth]{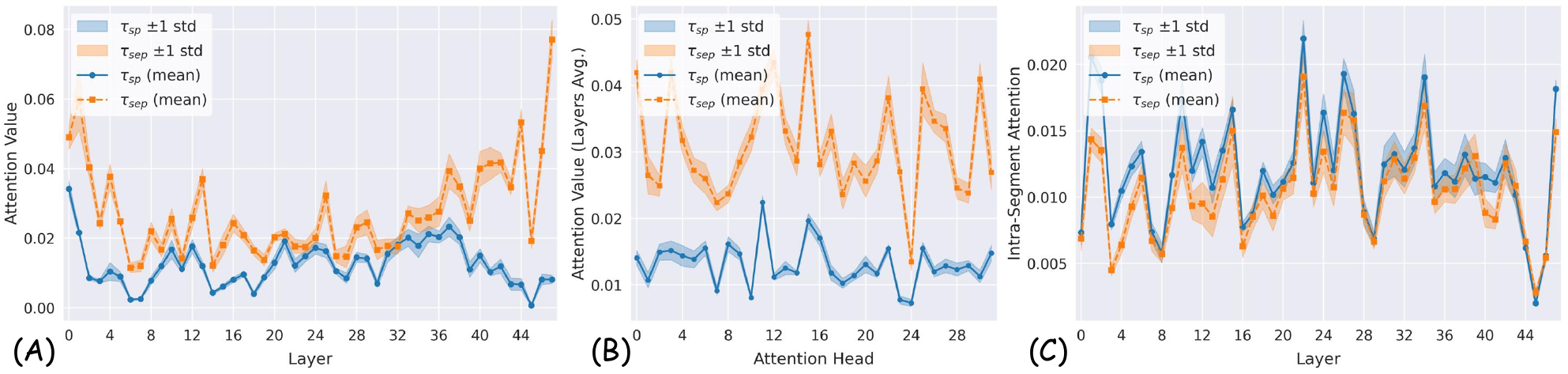}
    \caption{Visualization of the Separator Attention effect. (A-B) The separator $\tau_{sep}$ serves as a prominent attention sink, demonstrating consistently higher attention scores than $\tau_{sp}$ across both (A) model layers and (B) attention heads. (C) This concentrated attention on $\tau_{sep}$ results in a substantial reduction of cross-segment attention weights, thereby localizing the attention scope relative to the vanilla $\tau_{sp}$.}
    \label{fig:sep_attention}
\end{figure*}


Inspired by \citet{SepLLM}, which demonstrates that LLMs exhibit attention concentration on certain special tokens, \eg start/end markers in sequences, punctuation marks in sentences, and other separators. Furthermore, the semantic embedding vectors of these separators often encapsulate key information from their preceding segments.
Based on these observations, we propose SepSeq that guides LLMs to focus attention on local segments rather than the global sequence by artificially introducing specific separators into sequences.

Let $\mathcal{S} = [x_1, \dots, x_n]$ be a numerical sequence. In the vanilla baseline, a standard delimiter token $\tau_{sp}$ (\eg whitespace, comma) is inserted between adjacent numbers to distinguish them. The proposed SepSeq method imposes structure by replacing $\tau_{sp}$ with a specific separator token $\tau_{sep}$ (\eg `\texttt{\textbackslash n}') at every $k$-th interval. We formalize this as:
\begin{equation} 
\begin{aligned} 
    \text{Vanilla}(\mathcal{S}) = x_1 \oplus \tau_{sp} \oplus x_2 \oplus \tau_{sp} \oplus & \cdots \oplus x_n, \\
    \text{SepSeq}(\mathcal{S}, k) = \underbrace{x_1 \oplus \tau_{sp} \oplus \cdots \oplus x_k}_{\text{Segment } 1} \oplus & \tau_{sep} \\
     \oplus \underbrace{x_{k+1} \oplus \tau_{sp} \oplus \cdots \oplus x_{2k}}_{\text{Segment } 2} \oplus & \tau_{sep} \oplus \cdots
\end{aligned} 
\end{equation}
where $\oplus$ denotes concatenation and $k$ is the segment size. Crucially, this substitution strategy ensures that SepSeq maintains the identical token length as the vanilla baseline, avoiding any overhead in sequence length.
We mechanistically analyze how separator tokens alter the attention through the following assumption and theorem.

\begin{assumption}[Separator Attention]
    \label{assumption:sep_attention}
    Drawing upon observations from \citet{SepLLM}, we posit that the separator token $\tau_{sep}$ \textbf{attracts significantly more attention} from the subsequent context compared to standard delimiters $\tau_{sp}$.
    Formally, for a token $x_i$ appearing after the separator, we assume
    \begin{equation}
        A[i,\tau_{sep}] > A[i,\tau_{sp}],
    \end{equation}
    where $A[i,t]$ denotes the normalized attention weight from position $i$ to token $t$.
\end{assumption}

To empirically verify this, we analyze the attention affinity directed from subsequent context towards separator tokens versus standard delimiters. 
Specifically, we randomly sampled prompts to generate numerical sequences under both Vanilla and SepSeq formats.
We then computed the mean attention scores and variances across 10 independent trials, as detailed in Appendix~\ref{apd:separator_attention} and visualized in Figure~\ref{fig:sep_attention}(A,B).
The results reveal that separator token $\tau_{sep}$ consistently elicits significantly higher attention than standard delimiter $\tau_{sp}$ at both the layer and head levels, thereby providing strong empirical evidence corroborating Assumption~\ref{assumption:sep_attention}.

\begin{theorem}[Cross-Segment Attention Suppression]
\label{thm:attention_wall}
Under Assumption~\ref{assumption:sep_attention}, the separator token $\tau_{sep}$ suppresses the attention allocated to tokens in the preceding segment.
Formally, for a token $x_i$ appearing after $\tau_{sep}$ and a token $x_j$ appearing before $\tau_{sep}$,
\begin{equation}
    A_{\text{SepSeq}}[i,j] < A_{\text{Vanilla}}[i,j].
\end{equation}
\end{theorem}

We analyze the mechanistic role of separator tokens in modulating cross-segment attention patterns.
For vanilla method, given a sequence $\mathcal{S}_{vanilla} = [ \ldots, x_{j}, \tau_{sp}, x_{i}, \ldots]$, the attention score between positions $i$ and $j$ is computed using query vector $\mathbf{Q}_i$, key vector $\mathbf{K}_j$, and key dimension $d_k$:
\begin{equation} 
    A_{\text{vanilla}}[i,j] = \frac{\exp\left(\mathbf{Q}_i \cdot \mathbf{K}_j^T/\sqrt{d_k}\right)}{\sum_{l\leq i} \exp\left(\mathbf{Q}_i \cdot \mathbf{K}_l^T/\sqrt{d_k}\right)}. 
\end{equation}
In our SepSeq framework, with the sequence $\mathcal{S}_{SepSeq} = [\ldots, x_{j}, \tau_{sep}, x_{i},\ldots ]$, the attention calculation remains structurally identical:
\begin{equation}
A_{\text{SepSeq}}[i,j] = \frac{\exp\left(\mathbf{Q}_i \cdot \mathbf{K}_j^T/\sqrt{d_k}\right)}{\sum_{l\leq i} \exp\left(\mathbf{Q}_i \cdot \mathbf{K}_l^T/\sqrt{d_k}\right)}.
\end{equation}
However, strictly distinguishable behaviors emerge in the denominator. 
As illustrated in Figure~\ref{fig:sep_attention}, $\tau_{sep}$ functions as a prominent attention sink, eliciting significantly higher attention scores than standard delimiters $\tau_{sp}$. Specifically, for a token $x_{i}$ succeeding $\tau_{sep}$, the term $\mathbf{Q}_{i}\cdot \mathbf{K}_{sep}^T$ is substantial, whereas the baseline counterpart $\mathbf{Q}_{i}\cdot \mathbf{K}_{sp}^T$ is negligible.
This dominance inflates the denominator of the Softmax operation.
Consequently, given that the denominator term $Z^{\text{SepSeq}} > Z^{\text{Vanilla}}$, the attention assigned to any distant cross-segment token $x_j$ is asymptotically attenuated:
\begin{equation}
\frac{A_{\text{SepSeq}}[x_{i},x_{j}]}{A_{\text{Vanilla}}[x_{i},x_{j}]} < 1.
\end{equation}
This theoretical derivation is empirically corroborated in Figure~\ref{fig:sep_attention}(C) (see Appendix~\ref{apd:separator_attention} for details).



\section{Experiments}
To evaluate our approach, we structure our experiments around three core research questions (RQs) regarding performance, robustness, and efficiency:

\textbf{RQ1 -- Effectiveness and Generalization.} Does our proposed method consistently enhance model performance across diverse tasks and architectures, demonstrating strong generalizability?

\textbf{RQ2 -- Robustness and Sensitivity.} Which factors modulate the effectiveness of our method, and how can it be optimally configured for different scenarios?

\textbf{RQ3 -- Efficiency and Cost.} Does our method reduce token consumption during inference?

The remainder of this section is organized as follows: We first describe the experimental setup, including datasets, evaluation metrics, and baselines. We then provide a detailed analysis of the experimental results corresponding to each RQ.

\subsection{Experimental Settings}
\subsubsection{Dataset}\label{sec:datasets}
We design two datasets to conduct an in-depth investigation of LLMs' capabilities for processing long numerical sequences: a synthetic dataset $\mathcal{D}_\text{syn}$ and a real dataset $\mathcal{D}_\text{real}$. 

For $\mathcal{D}_\text{syn}$, we construct sequences of varying lengths comprising both integer and floating-point numbers. We categorize these sequences into four length intervals: S (short) for sequences containing [2, 32] (\ie $2\leq n\leq 32$) numbers, M (medium) for (32, 128] numbers, L (large) for (128, 256] numbers, and XL (extra-large) for (256, 512] numbers.
We formulate six distinct task types: (1) \emph{max-int}, which requires identifying the index of the maximum integer in an integer sequence; (2) \emph{min-int}, which locates the index of the minimum integer; (3) \emph{max-float} and (4) \emph{min-float}, which perform analogous operations on floating-point sequences; (5) \emph{indexing}, which determines the position of the last occurrence of 1 in a binary sequence; and (6) \emph{counting}, which counts the total number of 1s in a binary sequence. 
Each task comprises 200 samples, with 50 samples distributed across each of the four length categories.

Building upon the prior work of \citet{li2025numeric_bench}, we construct $\mathcal{D}_\text{real}$ to assess model performance on practical numerical reasoning tasks. This dataset comprises four distinct categories, each containing 200 samples. 
The categories include: (1) \emph{number-string}, which involves counting numerals in alphanumeric sequences; (2) \emph{number-list}, requiring logical reasoning over numerical sequences; and (3-4) \emph{stock} and \emph{weather}, both constructed from real-world datasets with human-generated questions. 
Notably, any failure to process a single value in these tasks inevitably results in an incorrect final answer, making them particularly challenging. See Appendix~\ref{apd:dataset_detail} for details (\eg examples, statistical information).

\subsubsection{Evaluation Metrics}
\label{sec:results_by_sep_symbol}

To comprehensively evaluate our proposed method, we assess both performance and robustness using the following metrics:

\textbf{Accuracy (Acc).} This metric serves as the primary indicator of performance, quantifying the ratio of correct predictions to the total number of test instances. Specifically, let $N_{\text{correct}}$ be the count of correct answers and $N$ be the total number of questions; Accuracy is defined as:
\begin{equation}
\text{Accuracy} = \frac{N_{\text{correct}}}{N}.
\end{equation}

\textbf{Answer Rate (AR).} 
Answer Rate measures reliability as the ratio of valid (non-null) responses to total inputs. Let $N$ be the total queries and $N_{\text{valid}}$ the count of successful generations, defined as:
\begin{equation}
    \text{AR} = \frac{N_{\text{valid}}}{N}.
\end{equation}

\subsubsection{Base Models}
We conduct a comprehensive evaluation across 9 LLMs, representing diverse architectures, parameter scales, and training paradigms from both open-source and proprietary domains.
\textbf{Open-source models:} Our selection includes the Qwen3 family~\citep{bg_reasoning_yang2025qwen3}, spanning 0.6B to 30B parameters with both dense and Mixture-of-Experts architectures~\citep{fedus2022moe,zhou2022moe-survey}, available in instruct and reasoning modes, alongside the QwQ-32B model. We also evaluate the DeepSeek series~\citep{bg_reasoning_guo2025deepseek_r1}, including the recent DeepSeek-R1 and DeepSeek-V3 variants, which are known for their strong reasoning capabilities.
\textbf{Proprietary models:} We assess Claude-3.7-Sonnet from Anthropic~\citep{bg_llm_claude}. Additionally, we evaluate Google's Gemini-2.5-Pro~\citep{bg_llm_gemini2.5pro}, which showcases multimodal understanding capabilities, and two variants from OpenAI's GPT-4 series~\citep{bg_reasoning_achiam2023gpt4}: GPT-4.1 and GPT-4o. See Section~\ref{sec:reproducibility} for a detailed version of the models.

\subsubsection{Baselines}
To rigorously evaluate the efficacy of our framework, we compare it against established inference paradigms.
\textbf{Vanilla:} As proposed by \citet{transformers_need_glasses}, we utilize standard delimiters (\eg comma) to separate numerical entries. This serves as a fundamental baseline, providing basic visual separation to enhance the representational distinctiveness of adjacent numbers in long sequences.
\textbf{Reasoning-Enhanced Inference:} We adopt Chain-of-Thought (CoT) prompting~\citep{cot} to activate the model's sequential reasoning. By explicitly prompting for a step-by-step derivation, this method enforces the decomposition of the problem into intermediate deductive steps.
\textbf{In-Context Learning:} Leveraging the in-context learning (ICL) capability of LLMs~\citep{one-shot}, this baseline provides a single exemplar within the prompt context. This setup tests the model's ability to induce the task format and logical pattern from sparse supervision.

\subsection{Experimental Results}
We present a comprehensive evaluation designed to address the proposed Research Questions (RQs). Each RQ is investigated through multiple complementary analytical lenses, providing rigorous empirical evidence to substantiate our claims.

\begin{table*}[!t]
\centering
\caption{The average answer rate and accuracy for each task over 10 independent runs, reported as mean $\pm$ standard deviation. \textbf{Bold} indicates the best average performance, while \underline{underlined} values denote the best average performance excluding our method. ``Incr.'' indicates the percentage improvement of SepSeq over the strongest non-SepSeq baseline. \textcolor{+}{Green} and \textcolor{-}{red} indicate improvement and degradation in performance, respectively.}
\label{tab:results-by-task-methods}
\small
\setlength{\tabcolsep}{1.4pt}
\begin{tabular}{l cccc cccc}
\toprule
\multirow{2.5}{*}{Task} & \multicolumn{4}{c}{Answer Rate (\%, $\uparrow$)} & \multicolumn{4}{c}{Accuracy (\%, $\uparrow$)} \\ 
\cmidrule(lr){2-5} \cmidrule(lr){6-9}
& Vanilla & CoT & ICL & SepSeq (Incr.) & Vanilla & CoT & ICL & SepSeq (Incr.)\\ 
\midrule

counting
& 100.0$\pm$0.0 & 100.0$\pm$0.0 & 100.0$\pm$0.0 & 100.0$\pm$0.0 (\phantom{0}+0.0\%)
& \underline{42.7$\pm$2.7} & 41.2$\pm$3.0 & 37.9$\pm$3.2 & \textbf{76.7$\pm$1.2} (\phantom{0}\textcolor{+}{+79.6\%}) \\

indexing
& 100.0$\pm$0.0 & 100.0$\pm$0.0 & 100.0$\pm$0.0 & 100.0$\pm$0.0 (\phantom{0}+0.0\%)
& \underline{38.8$\pm$3.1} & 33.4$\pm$3.4 & 34.9$\pm$2.9 & \textbf{86.6$\pm$0.8} (\textcolor{+}{+123.0\%}) \\

max-float
& 100.0$\pm$0.0 & 100.0$\pm$0.0 & 100.0$\pm$0.0 & 100.0$\pm$0.0 (\phantom{0}+0.0\%)
& \underline{63.9$\pm$1.9} & 60.5$\pm$2.1 & 63.3$\pm$1.7 & \textbf{81.0$\pm$1.1} (\phantom{0}\textcolor{+}{+26.8\%}) \\

max-int
& 100.0$\pm$0.0 & 100.0$\pm$0.0 & 100.0$\pm$0.0 & 100.0$\pm$0.0 (\phantom{0}+0.0\%)
& \underline{75.3$\pm$1.1} & 68.8$\pm$1.7 & 71.0$\pm$1.4 & \textbf{92.4$\pm$0.4} (\phantom{0}\textcolor{+}{+22.7\%}) \\

min-float
& 100.0$\pm$0.0 & 100.0$\pm$0.0 & 100.0$\pm$0.0 & 100.0$\pm$0.0 (\phantom{0}+0.0\%)
& \underline{63.1$\pm$1.7} & 60.2$\pm$2.2 & 60.4$\pm$1.9 & \textbf{79.3$\pm$1.0} (\phantom{0}\textcolor{+}{+25.7\%}) \\

min-int
& 100.0$\pm$0.0 & 100.0$\pm$0.0 & 100.0$\pm$0.0 & 100.0$\pm$0.0 (\phantom{0}+0.0\%)
& \underline{74.3$\pm$1.4} & 68.1$\pm$1.5 & 66.5$\pm$1.8 & \textbf{93.0$\pm$0.3} (\phantom{0}\textcolor{+}{+25.1\%}) \\

number-string
& \phantom{0}\underline{96.6$\pm$0.2} & \phantom{0}96.3$\pm$0.3 & \phantom{0}96.3$\pm$0.1 & \phantom{0}\textbf{99.0$\pm$0.1} (\phantom{0}\textcolor{+}{+2.5\%})
& 81.6$\pm$1.0 & 81.0$\pm$0.8 & \textbf{81.7$\pm$1.1} & \textbf{81.7$\pm$0.9} (\phantom{00}+0.0\%) \\

number-list
& \phantom{0}\underline{78.2$\pm$1.2} & \phantom{0}77.6$\pm$0.9 & \phantom{0}73.4$\pm$1.4 & \phantom{0}\textbf{79.1$\pm$1.1} (\phantom{0}\textcolor{+}{+1.1\%})
& \underline{36.3$\pm$3.3} & 35.4$\pm$3.1 & 32.6$\pm$3.5 & \textbf{36.7$\pm$3.2} (\phantom{00}\textcolor{+}{+0.9\%}) \\

stock
& \phantom{0}\underline{64.1$\pm$1.7} & \phantom{0}63.1$\pm$1.9 & \phantom{0}63.3$\pm$1.6 & \phantom{0}\textbf{73.6$\pm$1.4} (\textcolor{+}{+14.7\%})
& 13.2$\pm$4.4 & \underline{13.7$\pm$4.2} & 13.4$\pm$4.5 & \textbf{27.4$\pm$3.5} (\textcolor{+}{+100.0\%}) \\

weather
& \phantom{0}\underline{66.0$\pm$1.6} & \phantom{0}65.6$\pm$1.8 & \phantom{0}65.9$\pm$1.5 & \phantom{0}\textbf{76.8$\pm$1.2} (\textcolor{+}{+16.3\%})
& 26.7$\pm$3.8 & 27.2$\pm$3.5 & \underline{27.8$\pm$3.9} & \textbf{44.6$\pm$2.8} (\phantom{0}\textcolor{+}{+60.9\%}) \\

\midrule
Average
& \phantom{0}\underline{90.5$\pm$0.4} & \phantom{0}90.3$\pm$0.6 & \phantom{0}89.9$\pm$0.5 & \phantom{0}\textbf{92.8$\pm$0.3} (\phantom{0}\textcolor{+}{+2.6\%})
& \underline{51.6$\pm$2.5} & 49.0$\pm$2.7 & 48.9$\pm$2.4 & \textbf{69.9$\pm$1.6} (\phantom{0}\textcolor{+}{+35.6\%}) \\

\bottomrule
\end{tabular}
\end{table*}

\begin{table*}[!t]
\centering
\caption{The average answer rate and accuracy of each model across 10 tasks, computed over 10 independent runs and reported as mean $\pm$ standard deviation. \textbf{Bold} indicates the best average performance, while \underline{underlined} values denote the best average performance excluding our method. ``Incr.'' indicates the percentage improvement of SepSeq over the strongest non-SepSeq baseline. \textcolor{+}{Green} and \textcolor{-}{red} indicate improvement and degradation in performance, respectively.}
\label{tab:results_by_model_methods}
\small
\setlength{\tabcolsep}{1.2pt}
\begin{tabular}{l cccc cccc}
\toprule
\multirow{2.5}{*}{Model} & \multicolumn{4}{c}{Answer Rate (\%, $\uparrow$)} & \multicolumn{4}{c}{Accuracy (\%, $\uparrow$)} \\ 
\cmidrule(lr){2-5} \cmidrule(lr){6-9} 
& Vanilla & CoT & ICL & SepSeq (Incr.) & Vanilla & CoT & ICL & SepSeq (Incr.) \\ 
\midrule
Qwen3-8B & \phantom{0}79.4$\pm$1.1 & \phantom{0}\underline{79.5$\pm$0.9} & \phantom{0}75.2$\pm$1.3 & \phantom{0}\textbf{87.3$\pm$0.6} (\textcolor{+}{\phantom{0}+9.8\%}) & \underline{45.5$\pm$2.8} & 40.3$\pm$2.9 & 38.5$\pm$3.2 & \textbf{69.6$\pm$1.6} (\textcolor{+}{+53.0\%}) \\ 

Qwen3-30B-A3B & \phantom{0}80.6$\pm$0.9 & \phantom{0}80.9$\pm$1.1 & \phantom{0}\underline{81.0$\pm$0.8} & \phantom{0}\textbf{90.2$\pm$0.5} (\textcolor{+}{+11.4\%}) & \underline{54.3$\pm$2.4} & 52.7$\pm$2.2 & 52.3$\pm$2.5 & \textbf{80.7$\pm$1.1} (\textcolor{+}{+48.6\%}) \\ 

QwQ-32B & \phantom{0}\underline{72.7$\pm$1.5} & \phantom{0}72.2$\pm$1.3 & \phantom{0}\underline{72.7$\pm$1.2} & \phantom{0}\textbf{75.6$\pm$1.1} (\textcolor{+}{\phantom{0}+4.0\%}) & \underline{34.2$\pm$3.4} & 28.9$\pm$3.7 & 28.5$\pm$3.5 & \textbf{57.8$\pm$2.0} (\textcolor{+}{+69.0\%}) \\ 

DeepSeek-V3 & 100.0$\pm$0.0 & 100.0$\pm$0.0 & 100.0$\pm$0.0 & 100.0$\pm$0.0 ({\phantom{0}+0.0\%}) & \underline{45.7$\pm$2.6} & 44.5$\pm$2.9 & 45.2$\pm$2.7 & \textbf{50.9$\pm$2.5} (\textcolor{+}{+11.4\%}) \\ 

DeepSeek-R1 & \phantom{0}\textbf{99.9$\pm$0.1} & \phantom{0}\textbf{99.9$\pm$0.0} & \phantom{0}99.8$\pm$0.1 & \phantom{0}\textbf{99.9$\pm$0.0} ({\phantom{0}+0.0\%}) & \underline{61.1$\pm$1.8} & 56.6$\pm$2.3 & 57.9$\pm$2.0 & \textbf{70.5$\pm$1.4} (\textcolor{+}{+15.4\%}) \\ 

Claude-3.7-Sonnet & \phantom{0}99.8$\pm$0.1 & \phantom{0}\underline{99.9$\pm$0.0} & \textbf{100.0$\pm$0.0} & \phantom{0}\underline{99.9$\pm$0.1} (\textcolor{+}{\phantom{0}-0.1\%}) & 57.3$\pm$2.2 & \underline{57.6$\pm$2.0} & 54.2$\pm$2.4 & \textbf{79.1$\pm$1.1} (\textcolor{+}{+37.3\%}) \\ 

Gemini-2.5-Pro & \phantom{0}\underline{83.4$\pm$0.9} & \phantom{0}81.8$\pm$1.0 & \phantom{0}82.0$\pm$0.8 & \phantom{0}\textbf{84.6$\pm$0.7} (\textcolor{+}{\phantom{0}+1.4\%}) & \underline{58.9$\pm$2.2} & 48.0$\pm$2.7 & 56.9$\pm$2.0 & \textbf{79.1$\pm$1.0} (\textcolor{+}{+34.3\%}) \\ 

GPT-4.1 & \phantom{0}\underline{99.6$\pm$0.1} & \phantom{0}\underline{99.6$\pm$0.0} & \phantom{0}\textbf{99.8$\pm$0.1} & \phantom{0}99.5$\pm$0.1 (\textcolor{-}{\phantom{0}-0.1\%}) & \underline{61.0$\pm$1.9} & 60.0$\pm$2.1 & 60.0$\pm$1.8 & \textbf{82.7$\pm$0.9} (\textcolor{+}{+35.6\%}) \\ 

GPT-4o & \phantom{0}\textbf{99.0$\pm$0.1} & \phantom{0}98.5$\pm$0.2 & \phantom{0}\underline{98.6$\pm$0.1} & \phantom{0}\underline{98.6$\pm$0.1} (\textcolor{-}{\phantom{0}-0.4\%}) & 46.4$\pm$2.6 & \underline{51.5$\pm$2.5} & 44.5$\pm$2.9 & \textbf{59.1$\pm$2.1} (\textcolor{+}{+14.8\%}) \\ 

\midrule
Average & \phantom{0}\underline{90.5$\pm$0.5} & \phantom{0}90.3$\pm$0.4 & \phantom{0}89.9$\pm$0.6 & \phantom{0}\textbf{92.8$\pm$0.3} (\textcolor{+}{\phantom{0}+2.6\%}) & \underline{51.6$\pm$2.4} & 48.9$\pm$2.7 & 48.7$\pm$2.5 & \textbf{68.9$\pm$1.5} (\textcolor{+}{+35.6\%}) \\ 

\bottomrule
\end{tabular}
\end{table*}

\subsubsection{Effectiveness and Generalization (RQ1)}
\paragraph{Across tasks.}
We evaluate our method against baselines on six synthetic and four real tasks, reporting Accuracy and Answer Rate. Analysis in Table~\ref{tab:results-by-task-methods} reveals that standard LLMs struggle with long numerical sequences. Notably, methods like CoT and ICL prompting prove detrimental, dropping average accuracy from 51.6\% to 49.0\% and 48.9\%, respectively. This strongly suggests that LLM's failures stem not from an insufficient reasoning ability but from a fundamental inability to parse and manage numerical sequences properly.

Conversely, SepSeq achieves a 35.6\% relative accuracy gain (averaging 69.9\%) over the baseline. The improvement is particularly pronounced in real-world tasks, with surges of +100.0\% on Stock and +60.9\% on Weather datasets. Furthermore, SepSeq improves inference reliability (+2.6\% answer rate), demonstrating that simple input restructuring effectively rectifies sequence understanding failures without model retraining.

\paragraph{Across models.}
We evaluate 9 diverse high-performance LLMs, presenting model-wise performance gains in Table~\ref{tab:results_by_model_methods} to validate the general applicability of our approach. Our analysis demonstrates that SepSeq yields robust improvements averaging +35.6\% accuracy, irrespective of model family or architectural design.
\textbf{Model Families.} SepSeq proves effective across a wide spectrum of model lineages. It delivers consistent gains for proprietary state-of-the-art models, including GPT-4.1 (+35.6\%), Claude-3.7-Sonnet (+37.3\%), and Gemini-2.5-Pro (+34.3\%). Similarly, open-weights models from the Qwen and DeepSeek families exhibit comparable or even larger surges (e.g., QwQ-32B +69.0\%), confirming that the method's efficacy is not tied to a specific training recipe or alignment strategy.
\textbf{Architectures (Dense vs. MoE).} Crucially, the improvements extend across distinct architectural paradigms. For standard Dense models like Qwen3-8B, SepSeq unlocks a massive +53.0\% accuracy boost. The framework is equally potent for MoE architectures, such as Qwen3-30B-A3B, improving accuracy by 48.6\%.

\begin{figure*}[!t] 
    \centering
    \includegraphics[width=1.0\textwidth]{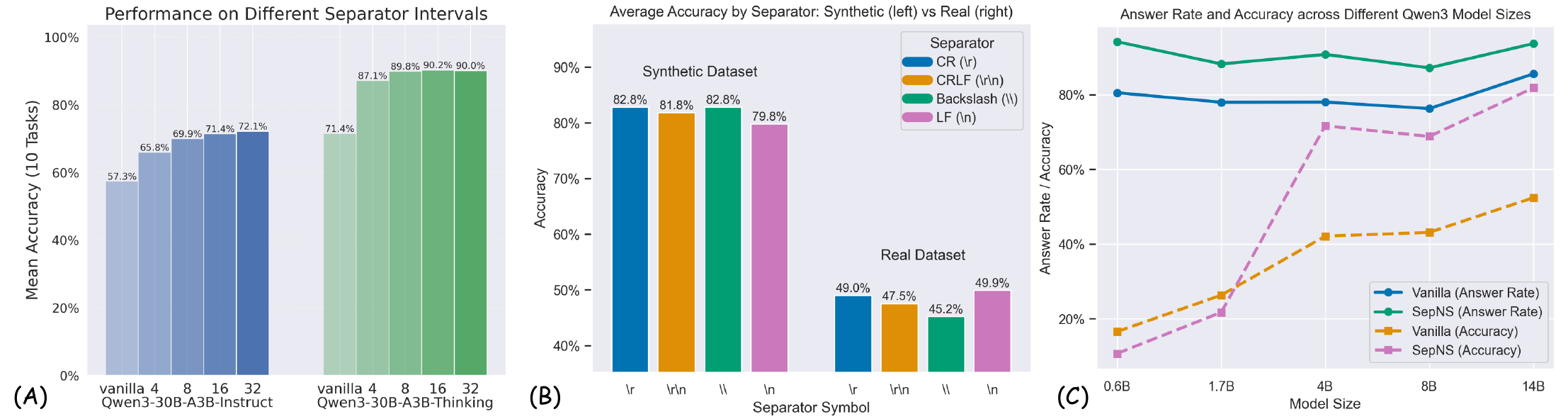}
    \caption{(A): Average Accuracy across different separate intervals. (B): Average Accuracy across different separator symbols. (C): Average Answer Rate and Accuracy across different Qwen3 model sizes: 0.6B to 14B.}
    \label{fig:model_size_sep_interval}
\end{figure*}

\subsection{Robustness and Sensitivity (RQ2)}

\paragraph{Separate interval.}

An analysis of the separator interval ($k$) reveals that our method exhibits remarkable stability once a minimal segmentation frequency is met. As shown in Figure~\ref{fig:model_size_sep_interval}(A), performance gains saturate rapidly: accuracy plateaus significantly for intervals $k \ge 8$, showing negligible variance for larger windows ($k=16, 32$). Specifically, the Thinking model maintains a consistent near-optimal accuracy of $\sim90\%$ across $k \in \{8, 16, 32\}$, with deviations of less than 0.4\%. Similarly, the Instruct model stabilizes in the 70-72\% range beyond this threshold. This observation suggests that SepSeq is hyperparameter-robust, providing substantial benefits without necessitating precise tuning of the segment length. We use $k=16$ in our main experiments. Based on these results, we set $k=16$ for our main experiments.

\paragraph{Separator symbol.}
We investigate separator symbol impact on performance using Qwen3-30B-A3B-Instruct-2507 across 10 tasks with four separator types: Carriage Return (CR, \texttt{\textbackslash r}), Carriage Return Line Feed (CRLF, \texttt{\textbackslash r\textbackslash n}), Backslash (\texttt{\textbackslash\textbackslash}), and Line Feed (LF, \texttt{\textbackslash n}). Tasks are categorized into basic numerical processing ($\mathcal{D}_{\text{syn}}$) and complex applications ($\mathcal{D}_{\text{real}}$).

Figure~\ref{fig:model_size_sep_interval}(B) reveals systematic performance differentiation across separator types varying with task complexity. 
For basic tasks $\mathcal{D}_{\text{syn}}$, all separators showed modest accuracy variations (79.8\%--82.8\%), with unconventional separators CR and Backslash achieving optimal accuracy (82.8\% each), potentially due to novelty requiring enhanced attention.
However, inversion emerged in complex scenarios: while CR and Backslash excel in basic tasks, LF demonstrated superior performance (49.9\% vs. Backslash's 45.2\%) in complex tasks. This reversal suggests fundamental processing strategy shifts across complexity levels. These patterns evidence sophisticated cognitive resource allocation~\citep{Sweller1988,Sweller2011} in LLMs. Under low cognitive load, models allocate additional resources to separator adaptation, where novel separators benefit from enhanced attention. Under high cognitive load, models prioritize core semantic processing, favoring minimal-overhead separators. LF's superior complex performance reflects the prevalence of training data and processing efficiency, enabling more resources for task-specific reasoning over format adaptation.

\paragraph{Model size.}
We evaluated Qwen3 models at various parameter scales (0.6B, 1.7B, 4B, 8B, and 14B). The results, shown in Figure~\ref{fig:model_size_sep_interval}(C), demonstrate that the effectiveness of SepSeq is scale-dependent. For smaller models (0.6B), SepSeq underperforms the vanilla baseline, suggesting a minimum capacity requirement for effective separator interpretation. A critical inflection point occurs at the 4B parameter level, where SepSeq begins to yield performance gains. Beyond this scale, the performance gap widens substantially, reaching over 80\% accuracy on the 14B models.

\begin{figure}[!t] 
    \centering
    \includegraphics[width=1.0\columnwidth]{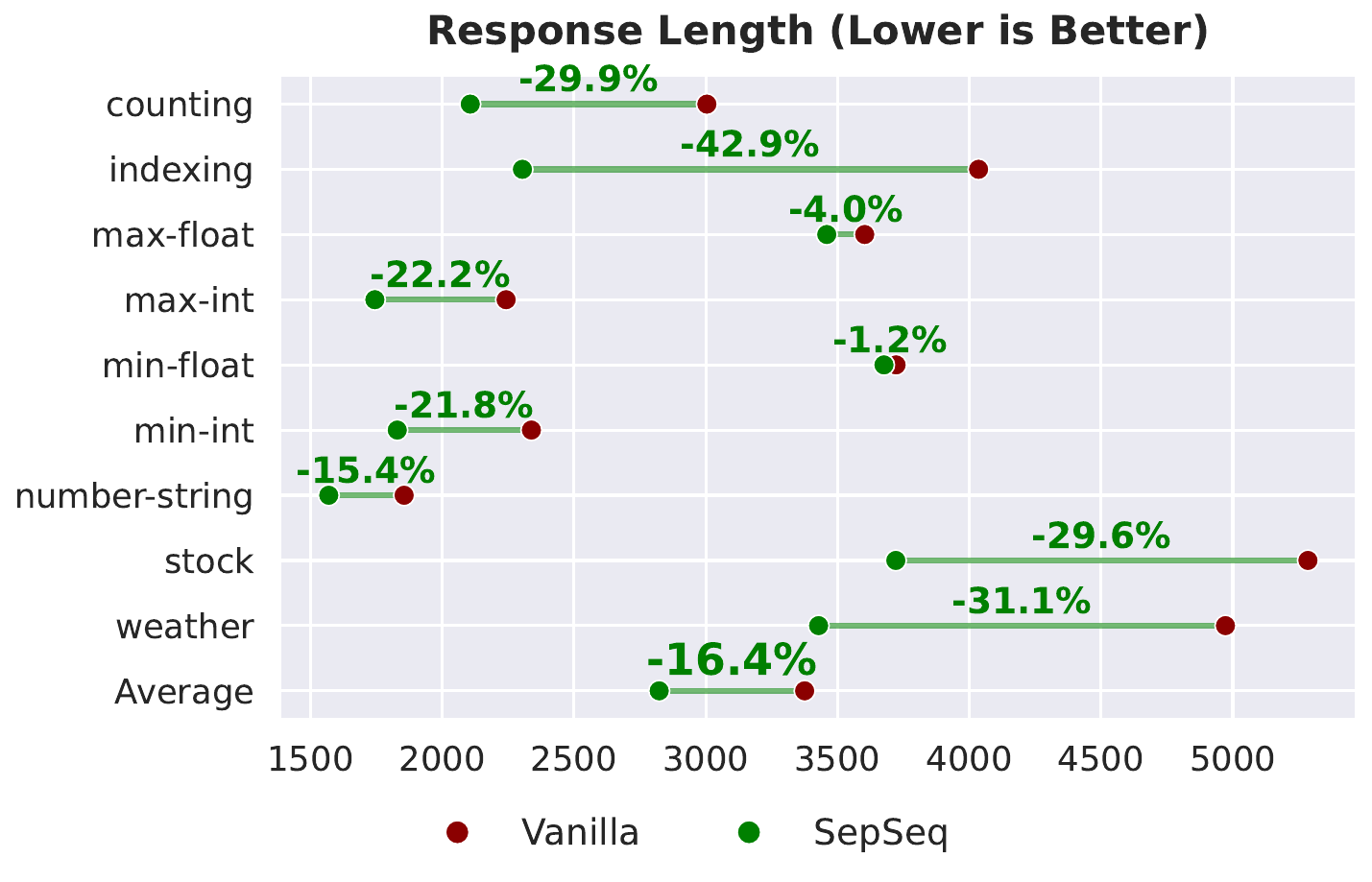}
    \caption{Comparison between \textcolor{-}{Vanilla} and \textcolor{+}{SepSeq} across tasks. The left panel shows that SepSeq significantly reduces token consumption during inference (average -16.4\%).}
    \label{fig:less_token}
\end{figure}

\paragraph{Reasoning vs. Instruction models.}
While SepSeq benefits different architectures, Figure~\ref{fig:model_size_sep_interval}(A) elicits a stronger response from reasoning-enhanced models. First, the Thinking model's baseline without separators (71.4\%) matches the Instruct model's saturated peak (72.1\%), indicating an intrinsic advantage in handling long contexts. Second, when augmented with SepSeq, the Thinking model achieves a massive leap to 90.2\% accuracy, breaking the performance ceiling that constrains the Instruct model. In contrast, the Instruct model, while improving from 57.3\% to 72.1\%. This disparity indicates that SepSeq acts as a structural multiplier for reasoning: it provides the necessary attention landmarks that reasoning models are uniquely equipped to exploit, unlocking superior numerical processing capabilities.


\paragraph{Comparison with Program-of-Thought.} 
\begin{table}[t]
\centering
\small
\setlength{\tabcolsep}{7pt}
\begin{tabular}{lcc}
\toprule
\multirow{2}{*}{Prompting} & \multicolumn{2}{c}{Input Formatting} \\
\cmidrule(lr){2-3}
& Standard & SepSeq \\
\midrule
Vanilla & 53.25 & 70.35 \\
PoT     & \underline{78.90} & \textbf{84.45} \\
\bottomrule
\end{tabular}
\caption{
Average results for PoT across 10 tasks (200 samples/task).
}
\label{tab:pot_2x2}
\end{table}
To clarify whether SepSeq replaces or complements tool-assisted reasoning, we compare it against a Program-of-Thought (PoT) baseline in a compact 2$\times$2 setting: \textit{Vanilla} vs.\ \textit{PoT}, each evaluated with and without SepSeq formatting. SepSeq is applied only to the input numerical sequence, while the PoT prompting and execution pipeline remain unchanged. As shown in Table~\ref{tab:pot_2x2}, across 10 tasks (200 samples per task), SepSeq consistently improves performance in both settings. Under standard prompting, accuracy increases from 53.25\% to 70.35\%; under PoT, it further improves from 78.90\% to 84.45\%. The gain persists even when combined with structured code-based reasoning, with PoT+SepSeq achieving the best overall result.

These findings indicate that SepSeq is complementary to PoT rather than redundant with it. While PoT primarily strengthens downstream computation, SepSeq improves the upstream processing of long numerical inputs, making the two approaches naturally compatible.

\subsection{Efficiency and Cost (RQ3)}

\paragraph{Token Consumption.}
We analyze the computational efficiency of SepSeq regarding token usage. One might worry that inserting separator tokens increases the total sequence length, thereby increasing inference cost. However, Figure~\ref{fig:less_token} presents a favorable outcome: SepSeq consistently reduces both response length and total token consumption across most tasks.

As shown in the chart, the \textit{Response Length} exhibits a drastic reduction. The average response length drops from 3,375 tokens (Vanilla) to 2,823 tokens (SepSeq), representing a 16.4\% reduction. In specific tasks like \textit{Indexing} and \textit{Stock}, the reduction is even more pronounced, reaching approximately 42.9\% and 29.6\%, respectively. This efficiency stems from SepSeq's ability to mitigate the ``repetition loops'' and hallucinated gibberish often observed in vanilla LLMs when they lose track of long sequences. By maintaining focused attention, SepSeq generates concise, precise answers without redundant tokens. This confirms that SepSeq is not only an accuracy-enhancing framework but also a cost-effective solution that reduces inference latency and API costs.
\section{Conclusion}

In this work, we identify and address a fundamental limitation of LLMs: attentional dispersion over long numerical sequences, which precipitates severe performance degradation in precision-critical tasks.
To mitigate this, we introduce SepSeq, a training-free framework that strategically inserts separators to partition sequences into manageable segments. 
Through evaluation across 9 state-of-the-art models and 10 diverse tasks, SepSeq achieves a substantial 35.6\% average accuracy improvement. 
Crucially, this gain comes without additional training and reduces net token consumption. 
Our analysis reveals that separators induce localized attention patterns, acting as anchors that transform dispersed attention into focused segment processing. 
This demonstrates that simple input restructuring serves as a powerful mechanism to unlock LLM numerical capabilities, offering a scalable and efficient solution for real-world applications.

\section*{Limitations}

While SepSeq demonstrates significant improvements in processing long numerical sequences, we identify the limitation in our current study:

\textbf{Scale Dependence.} Our evaluation on the Qwen3 family reveals that SepSeq's effectiveness is closely tied to model scale. Specifically, models with fewer parameters (\eg 0.6B) exhibit performance degradation compared to the vanilla baseline, likely due to their limited capacity to interpret the implicit structural guidance provided by separators. The performance gains only become robust significantly beyond the 4B parameter threshold, suggesting that our method requires a certain level of emergent cognitive capability to be effective.

\section*{Ethical Consideration} 
The proposed SepSeq framework is a training-free inference technique that modifies input formatting without altering model parameters or requiring additional data collection. 
Our experiments utilize publicly available models and synthetic datasets, with no involvement of human subjects, collection of personal data, or privacy risks. 
The method enhances model accuracy in numerical processing tasks without introducing harmful capabilities or creating potential for misuse. 
All experimental evaluations were conducted using established benchmarks and standard evaluation protocols. 
The research makes a positive contribution to the field by addressing fundamental limitations in LLM numerical processing capabilities, with potential benefits for applications that require precise numerical computation.



\bibliography{acl_latex}

\appendix


\section{The Long-Context Copying Bottleneck}
\label{apd:repetition-dilemma}
To evaluate the capability of LLMs on repetition, we conducted a preliminary experiment on ``strict numerical sequence repetition.'' The task requires a model to reproduce a given numerical sequence exactly, without any additions, omissions, or alterations. We designed a testing framework with progressively increasing difficulty by dividing sequence lengths into five ranges: 2--32, 33--128, 129--256, 257--512, and 513--1024. For each range, 50 unique samples were randomly generated. Each sequence consisted of numbers with three decimal places, drawn uniformly from the interval [-10, 10]. Models were prompted to return the output in a strict JSON array format (\eg [1.234, -5.678, 9.012]), prohibiting any extraneous characters, spaces, or line breaks. A response was judged as correct only if it was an exact string match to the ground truth sequence.

The experiment was performed on two variants of the Qwen3 model: Qwen3-30B-A3B-Instruct-2507 and Qwen3-30B-A3B-Thinking-2507. To ensure deterministic and stable outputs, the decoding temperature was set to 0. Any deviation in format or content from the expected output was classified as an error.

\begin{table*}[t]
\centering
\caption{Performance of Qwen3-30B-A3B-Instruct-2507 and Qwen3-30B-A3B-Thinking-2507 on the strict numerical sequence repetition task. The table shows the number of correct reproductions and accuracy for each sequence length range.}
\small
\label{tab:repetition_dilemma}
\begin{tabular}{@{}lrrrrrr@{}}
\toprule
\multirow{2.5}{*}{Size} & \multirow{2.5}{*}{Total} & \multicolumn{2}{c}{Qwen3-30B-A3B-Instruct Model} & \multicolumn{2}{c}{Qwen3-30B-A3B-Thinking-2507} \\
\cmidrule(lr){3-4} \cmidrule(lr){5-6} 
 & & \#~Correct & Accuracy & \#~Correct & Accuracy  \\
\midrule
S: 2-32 & 50 & 50 & 100.00\% & 50 & 100.00\% \\
M: 33-128 & 50 & 50 & 100.00\% & 36 & 72.00\% \\
L: 129-256 & 50 & 50 & 100.00\% & 16 & 32.00\% \\
XL: 257-512 & 50 & 7 & 14.00\% & 0 & 0.00\% \\
XXL: 513-1024 & 50 & 0 & 0.00\% & 0 & 0.00\% \\
\midrule
\textbf{Overall} & 250 & 157 & 62.80\% & 102 & 40.80\% \\
\bottomrule
\end{tabular}
\end{table*}

The results, presented in Table~\ref{tab:repetition_dilemma}, reveal a strong ``repetition dilemma'' in both models. The Qwen3-30B-A3B-Instruct-2507 variant performed flawlessly on sequences up to 256 numbers, achieving 100\% accuracy. However, its performance collapsed to just 14\% accuracy in the 257--512 range (XL) and failed completely on the longest sequences (XXL). Counter-intuitively, the Qwen3-30B-A3B-Thinking-2507 variant, despite its designation, demonstrated inferior overall performance (40.80\% vs. 62.80\%). Its accuracy began to degrade significantly on medium-length sequences (M and L), falling far short of its instruct-tuned counterpart.

These findings highlight significant architectural or attentional limitations in current LLMs for tasks demanding precise, long-sequence replication. Such failures may stem from compounding errors in the attention mechanism, effective context window constraints, or biases in the training data. The inferior performance of the ``Thinking'' variant is particularly noteworthy. It suggests that for rote, mechanical tasks that do not require reasoning, the cognitive overhead or architectural modifications intended to facilitate complex thought may act as a source of noise, thereby degrading performance on simple memorization and reproduction.

\section{Dataset Details}
\label{apd:dataset_detail}

This appendix provides detailed descriptions of the datasets used to evaluate LLMs' long numerical sequence processing capabilities.

\subsection{Synthetic Dataset}

The synthetic dataset $D_{\text{syn}}$ consists of 1,200 samples across six task types, each containing 200 samples. The sequences are categorized into four length intervals:

\begin{itemize}
    \item \textbf{S (Short)}: [2, 32] numbers (50 samples per task)
    \item \textbf{M (Medium)}: (32, 128] numbers (50 samples per task)  
    \item \textbf{L (Large)}: (128, 256] numbers (50 samples per task)
    \item \textbf{XL (Extra-large)}: (256, 512] numbers (50 samples per task)
\end{itemize}

\subsubsection{Task Types}

\textbf{max-int}: Identify the index (0-based) of the maximum integer in an integer sequence. Example:
\begin{lstlisting}[language=json]
{
  "task_type": "max_int", 
  "answer": "7", 
  "ts": [3, 2, 2, 0, 3, 0, 2, 5, 0, 0, 1, 0, 1, 0, 1, 2, 0, 1, 4]
}
\end{lstlisting}
The maximum value is 5 at index 7.

\noindent\textbf{min-int}: Identify the index (0-based) of the minimum integer in an integer sequence. Example:
\begin{lstlisting}[language=json]
{
  "task_type": "min_int", 
  "answer": "19", 
  "ts": [6, 9, 7, 6, 7, 7, 6, 6, 6, 7, 9, 8, 7, 6, 6, 8, 8, 8, 9, 2, 9, 9, 6, 9, 6, 8, 6, 9, 6]
}
\end{lstlisting}
The minimum value is 2 at index 19.

\noindent\textbf{max-float}: Identify the index (0-based) of the maximum floating-point number in a sequence. Similar to max-int but with floating-point numbers.

\noindent\textbf{min-float}: Identify the index (0-based) of the minimum floating-point number in a sequence. Similar to min-int but with floating-point numbers.

\noindent\textbf{indexing}: Determine the position of the last occurrence of 1 in a binary sequence. Example:
\begin{lstlisting}[language=json]
{
  "task_type": "indexing", 
  "answer": "8", 
  "ts": [1, 0, 0, 0, 1, 0, 0, 1, 1, 0, 0, 0, 0, 0, 0, 0, 0, 0, 0, 0, 0, 0]
}
\end{lstlisting}
The last occurrence of 1 is at index 8.

\noindent\textbf{counting}: Count the total number of 1s in a binary sequence. Example:
\begin{lstlisting}[language=json]
{
  "task_type": "counting", 
  "answer": "4", 
  "ts": [1, 0, 0, 0, 1, 0, 0, 1, 1, 0, 0, 0, 0, 0, 0, 0, 0, 0, 0, 0, 0, 0]
}
\end{lstlisting}
There are 4 occurrences of 1 in the sequence.

\subsection{Real Dataset}

The real dataset $D_{\text{real}}$ consists of 800 samples across four categories, each containing 200 samples. These tasks are based on practical numerical reasoning scenarios.

\subsubsection{Task Categories}

\textbf{number-string}: Count numerals in alphanumeric sequences. Example:
\begin{lstlisting}[language=json]
{
  "question": "How many numbers are there in the string? Note that a sequence like 'a243b' counts as a single number.",
  "struct_data": "effV2xM8hF5vcNgl8xrTCmbD6sEM38ti",
  "answer": 11
}
\end{lstlisting}
This task requires parsing mixed alphanumeric strings to identify and count distinct numerical sequences.

\noindent\textbf{number-list}: Perform logical reasoning over numerical sequences with multiple-choice questions. Example:
\begin{lstlisting}[language=json]
{
  "question": "Which index holds the greatest number in the list between the indices 20 and 80? Options: A: 40, B: 75, C: 53, D: 58, E: 48, F: 44, G: 60, H: 31",
  "struct_data": [1372.31, -3479.74, 1046, "..."],
  "answer": "H"
}
\end{lstlisting}
These tasks involve complex reasoning operations such as finding extrema within specific ranges, identifying patterns, or performing conditional operations.

\noindent\textbf{stock}: Answer questions about financial time series data. Example:
\begin{lstlisting}[language=json, numbers=none, caption={Stock example (truncated)}, label={lst:stock-example}]
{
  "question": "How many days had a volume over 15,000 between 2024-10-15 and 2024-10-25? Options: A: 3, B: 5, C: 7, D: 9",
  "struct_data": [
    {"date": "2024-10-15", "close_price": 52.56, "volume": 24421, "...": "..."},
    {"date": "2024-10-16", "close_price": 52.80, "volume": 19962, "...": "..."},
    {"date": "2024-10-17", "close_price": 53.11, "volume": 19210, "...": "..."},
    {"date": "2024-10-18", "close_price": 55.11, "volume": 25238, "...": "..."},
    "..."
  ],
  "answer": "C"
}
\end{lstlisting}
This task involves analyzing real-world financial time series data with questions about trading volumes, price movements, and temporal patterns.

\noindent\textbf{weather}: Answer questions about meteorological time series data. Example:
\begin{lstlisting}[language=json, numbers=none, caption={Weather example (truncated)}, label={lst:weather-example}]
{
  "question": "On which date was the temperature lastly above 5 degrees between 2024-11-10 and 2024-11-20? Options: A: 2024-11-11, B: 2024-11-14, C: 2024-11-12, D: 2024-11-13",
  "struct_data": [
    {"date": "2024-11-10", "temperature_2m": 5.99, "precipitation": 0.0, "...": "..."},
    {"date": "2024-11-11", "temperature_2m": 5.51, "precipitation": 0.0, "...": "..."},
    {"date": "2024-11-12", "temperature_2m": 5.10, "precipitation": 0.0, "...": "..."},
    {"date": "2024-11-13", "temperature_2m": 4.92, "precipitation": 0.0, "...": "..."},
    {"date": "2024-11-14", "temperature_2m": 5.45, "precipitation": 0.0, "...": "..."},
    "..."
  ],
  "answer": "B"
}
\end{lstlisting}
This task involves analyzing real-world meteorological time series data with questions about temperature patterns, precipitation, and temporal trends.

\subsection{Dataset Statistics}
Table~\ref{tab:detailed-token-stats} shows the dataset statistics.

\begin{table*}[!t]
\centering
\caption{Comprehensive Token Length Statistics. Comparison between Question Only and Full Prompt across six statistical dimensions: Minimum, Quartiles (25\%, 50\%, 75\%), Mean, and Maximum. All tasks consist of 200 samples.}
\label{tab:detailed-token-stats}
\small
\setlength{\tabcolsep}{3pt} 
\begin{tabular}{l rrrrrr rrrrrr c}
\toprule
\multirow{2.5}{*}{Task Type} & \multicolumn{6}{c}{Question Only (Tokens)} & \multicolumn{6}{c}{Full Prompt (Tokens)} & \multirow{2.5}{*}{\#Samples} \\ 
\cmidrule(lr){2-7} \cmidrule(lr){8-13}
& Min & 25\% & 50\% & Mean & 75\% & Max & Min & 25\% & 50\% & Mean & 75\% & Max & \\ 
\midrule
max\_int & 56 & 56 & 56 & 56.0 & 56 & 56 & 88 & 153 & 341 & 411.9 & 594 & 1070 & 200 \\ 

min\_int & 56 & 56 & 56 & 56.0 & 56 & 56 & 88 & 149 & 341 & 410.3 & 594 & 1104 & 200 \\ 

max\_float & 56 & 56 & 56 & 56.0 & 56 & 56 & 106 & 318 & 957 & 1252.8 & 1867 & 3613 & 200 \\ 

min\_float & 56 & 56 & 56 & 56.0 & 56 & 56 & 99 & 334 & 968 & 1232.3 & 1869 & 3562 & 200 \\ 

counting & 24 & 24 & 24 & 24.0 & 24 & 24 & 56 & 117 & 309 & 392.9 & 565 & 1076 & 200 \\ 

indexing & 43 & 43 & 43 & 43.0 & 43 & 43 & 75 & 138 & 328 & 408.8 & 588 & 1085 & 200 \\ 

weather & 66 & 92 & 134 & 119.0 & 140 & 145 & 9630 & 9817 & 9853 & 9859.3 & 9910 & 10127 & 200 \\ 

stock & 67 & 86 & 131 & 117.8 & 138 & 151 & 11024 & 11493 & 11674 & 11746.9 & 11992 & 12521 & 200 \\ 

number\_list & 55 & 83 & 100 & 116.1 & 162 & 194 & 640 & 772 & 5079 & 3916.1 & 5124 & 5195 & 200 \\ 

mixed\_number\_string & 27 & 27 & 27 & 27.0 & 27 & 27 & 128 & 135 & 137 & 136.7 & 139 & 141 & 200 \\ 

\bottomrule
\end{tabular}
\end{table*}

    


\section{Separator Attention Effect}
\label{apd:separator_attention}
\subsection{Separator Attention}
\textbf{Experimental Setup.} To quantify the attention distribution mechanism, we conducted a controlled experiment using the Qwen3-30B-A3B-Thinking model. We generated 10 independent trials of random numerical sequences, each consisting of 20 integers (range 0-9). The sequences were formatted using two configurations: \textbf{Vanilla:} Uses standard spaces ($\tau_{sp}$) as delimiters; \textbf{SepSeq:} Inserts newline separators ($\tau_{sep}$) every 8 tokens. 

\textbf{Metric Definition.} We extracted the attention weights from tokens in subsequent segments (specifically indices 9-16 and 17-20) directed towards the separator tokens located at the segment boundaries (indices corresponding to the 1st and 2nd separators). Figure~\ref{fig:sep_attention}(A) reports the attention scores averaged across all heads for each layer, with shaded regions representing $\pm 1$ standard deviation across the 10 sequences. Figure~\ref{fig:sep_attention}(B) reports the attention scores for each attention head index (0-31), averaged across all layers and sequences, to illustrate head-specific behaviors.

\subsection{Cross-Segment Attention Suppression}
\textbf{Experimental Setup for Cross-Segment Attention Analysis.} To investigate the impact of separator tokens on long-range dependency modeling, we analyzed the inter-segment attention patterns using the same Qwen3-30B-A3B-Thinking model and data generation protocol as the previous experiment (10 independent sequences of length 20). We compared two formatting strategies: \textbf{Vanilla:} Standard space-delimited sequence; \textbf{SepSeq:} Segmented sequence with newline separators inserted every 8 tokens.

\textbf{Metric Definition.} We defined Cross-Segment Attention as the aggregate attention weight originating from tokens in a current segment and directed towards tokens in all preceding segments. Specifically, we computed the attention flow for two cross-segment scenarios: (1) From tokens in the 2nd segment (indices 9-16) attending to tokens in the 1st segment (indices 1-8); (2)From tokens in the 3rd segment (indices 17-20) attending to tokens in the combined 1st and 2nd segments (indices 1-16).

\textbf{Visualization.} The Figure~\ref{fig:sep_attention}(C) reports the mean cross-segment attention scores averaged across all attention heads and selected token pairs for each model layer. Shaded areas indicate the standard deviation across the 10 independent trials. A lower value in the SepSeq configuration indicates effective suppression of attention to distant, irrelevant context.

\section{The Use of Large Language Models}

We employed LLMs for bug detection in code. Additionally, LLMs were utilized to refine and polish the manuscript content based on specific requirements. All LLM-generated content, including code and textual revisions, underwent thorough review and validation by the authors to ensure accuracy, quality, and alignment with our research objectives.

\section{Reproducibility statement}
\label{sec:reproducibility}

To facilitate reproducibility of our results, we provide comprehensive documentation and resources across multiple components of this work. 

\textbf{Code and Data.} Our proposed method is thoroughly detailed in Section~\ref{sec:method}, including algorithmic descriptions and implementation specifics. Complete source code and datasets are available through the anonymous repository at \url{https://anonymous.4open.science/r/SepSeq}, with the accompanying README file providing step-by-step instructions for execution and reproduction of experiments.

\textbf{Selected Models.} We evaluate diverse high-performance models across different experimental settings. For the \textbf{main evaluation (RQ1)}, we use 9 high-performance LLMs:
\begin{enumerate}[leftmargin=*]
    \item Qwen3-8B: \url{https://huggingface.co/Qwen/Qwen3-8B}
    \item Qwen3-30B-A3B: \url{https://huggingface.co/Qwen/Qwen3-30B-A3B}
    \item QwQ-32B: \url{https://huggingface.co/Qwen/QwQ-32B}
    \item DeepSeek-V3: \url{https://huggingface.co/deepseek-ai/DeepSeek-V3-0324}
    \item DeepSeek-R1: \url{https://huggingface.co/deepseek-ai/DeepSeek-R1-0528}
    \item Claude-3.7-Sonnet: \url{https://openrouter.ai/anthropic/claude-3.7-sonnet}
    \item Gemini-2.5-Pro: \url{https://openrouter.ai/google/gemini-2.5-pro}
    \item GPT-4.1: \url{https://openrouter.ai/openai/gpt-4.1}
    \item GPT-4o: \url{https://openrouter.ai/openai/gpt-4o-2024-08-06}
\end{enumerate}

For robustness evaluation (RQ2), \textbf{separator interval analysis}, we compare instruction and reasoning variants:
\begin{enumerate}[leftmargin=*]
    \item Qwen3-30B-A3B-Instruct: \url{https://huggingface.co/Qwen/Qwen3-30B-A3B-Instruct-2507}
    \item Qwen3-30B-A3B-Thinking: \url{https://huggingface.co/Qwen/Qwen3-30B-A3B-Thinking-2507}
\end{enumerate}

For \textbf{separator symbol analysis}, we use Qwen3-30B-A3B-Instruct (\url{https://huggingface.co/Qwen/Qwen3-30B-A3B-Instruct-2507}). 

For \textbf{model size analysis}, we evaluate across different parameter scales:
\begin{enumerate}[leftmargin=*]
    \item Qwen3-0.6B: \url{https://huggingface.co/Qwen/Qwen3-0.6B}
    \item Qwen3-1.7B: \url{https://huggingface.co/Qwen/Qwen3-1.7B}
    \item Qwen3-4B: \url{https://huggingface.co/Qwen/Qwen3-4B}
    \item Qwen3-8B: \url{https://huggingface.co/Qwen/Qwen3-8B}
    \item Qwen3-14B: \url{https://huggingface.co/Qwen/Qwen3-14B}
\end{enumerate}




\end{document}